# Ego-motion and Surrounding Vehicle State Estimation Using a Monocular Camera

Jun Hayakawa, Behzad Dariush

*Abstract*— Understanding ego-motion and surrounding vehicle state is essential to enable automated driving and advanced driving assistance technologies. Typical approaches to solve this problem use fusion of multiple sensors such as LiDAR, camera, and radar to recognize surrounding vehicle state, including position, velocity, and orientation. Such sensing modalities are overly complex and costly for production of personal use vehicles. In this paper, we propose a novel machine learning method to estimate ego-motion and surrounding vehicle state using a single monocular camera. Our approach is based on a combination of three deep neural networks to estimate the 3D vehicle bounding box, depth, and optical flow from a sequence of images. The main contribution of this paper is a new framework and algorithm that integrates these three networks in order to estimate the ego-motion and surrounding vehicle state. To realize more accurate 3D position estimation, we address ground plane correction in real-time. The efficacy of the proposed method is demonstrated through experimental evaluations that compare our results to ground truth data available from other sensors including Can-Bus and LiDAR.

## I. INTRODUCTION

Understanding ego-motion and surrounding vehicle state is one of the critical factors for successful deployment of autonomous and ADAS-enabled vehicles. The 3D position, velocity, and orientation of surrounding vehicles provide information that are critical for decision making and path planning strategies in automated vehicles. In addition, for autonomous technologies to be widely adopted in production vehicles, it is important for such systems to be as simple as possible to ease implementation and ensure reliability while minimizing cost.

There is an extensive body of research for understanding ego-motion and surrounding vehicle state. The performance of the resulting technologies has significantly improved with the emergence of algorithms that adopt deep neural networks. Most approaches rely on LiDAR [1] or fusion of multiple sensors [2][3] such as LiDAR, cameras, and radar to estimate the state of surrounding vehicles. Other approaches [4][5][8][9] use cameras exclusively to make the vehicle sensor system cost-effective and straightforward. Despite such advances in vision based methods, it is still challenging to estimate the surrounding vehicle state given the limited information provided from a monocular camera as compared to methods that rely on fusion of information from multiple sensors.

To realize camera based solutions for surround vehicle state estimation, core technologies for vehicle detection and depth estimation are required. Object detection is one such technology that is required to predict a surrounding vehicle state such as position, velocity, and orientation. Recently 2D object detection algorithms such as YOLO 9000 [6], SSD [19], and FPN FRCN [7] have achieved great performance with fast and accurate 2D object detection using a monocular camera. Although the 2D object detection methods provide detection and localization of the vehicle in the image, they lack distance information.

Since 3D object detection is necessary to establish detailed vehicle state, LiDAR is generally used in autonomous vehicle development to estimate 3D features of the vehicles [1]. However, LiDAR technology is still prohibitively expensive for use in production vehicles. Furthermore, LiDAR with rotating parts may not provide the long-term reliability required in automotive applications. Alternatively, stereo cameras can be used for estimating the distance to surrounding vehicles by obtaining relative depth information in the form of a disparity map. However, a stereo camera is also expensive and requires high precision calibration. Since a monocular camera provides a simple and low cost sensing modality, research in this area is important for 3D surround vehicle state estimation.

Recent advances in the area of 3D object detection using a monocular camera include, for example, estimation of 3D bounding box from detected 2D bounding box [5], and monocular 3D object detection using CNN [8]. These approaches are based on regression of the 3D bounding box in the 2D image. Another approach using a monocular camera proposed by Libor Novák [9] combines the state-of-the-art methods for 2D bounding box detection. Their approach can be optimized to regress the positions of 3D bounding boxes by reconstructing the 3D world using the ground plane. However, the reconstructed 3D world position of the detected bounding boxes is heavily dependent on estimation accuracy of the ground plane. Furthermore, this approach assumes a single flat constant ground plane and outputs the surrounding vehicle state such as 3D position and orientation. The assumption of a fixed single ground plane is particularly limiting in driving situations, especially when the road surface shape is uneven, or the driver suddenly accelerates or brakes.

Clearly, reliable ground plane estimation is an important core technology for accurate 3D position estimation of objects from an image sequence. The work in [13] addresses this problem based on the assumption that 3D ego-motion and the ground plane normal vector are orthogonal. They formulate the problem as a state-continuous Hidden Markov Model though homography decomposition. This technique is particularly useful in cases where direct measurement of the ground plane parameters are not available and real-time

Jun Hayakawa and Behzad Dariush are with Honda Research Institute, 375 Ravendale Dr., Suite B, Mountain View, CA 94043 USA. (Phone: +1-650-314-0400; e-mail: {jhayakawa, bdariush}@honda-ri.com).

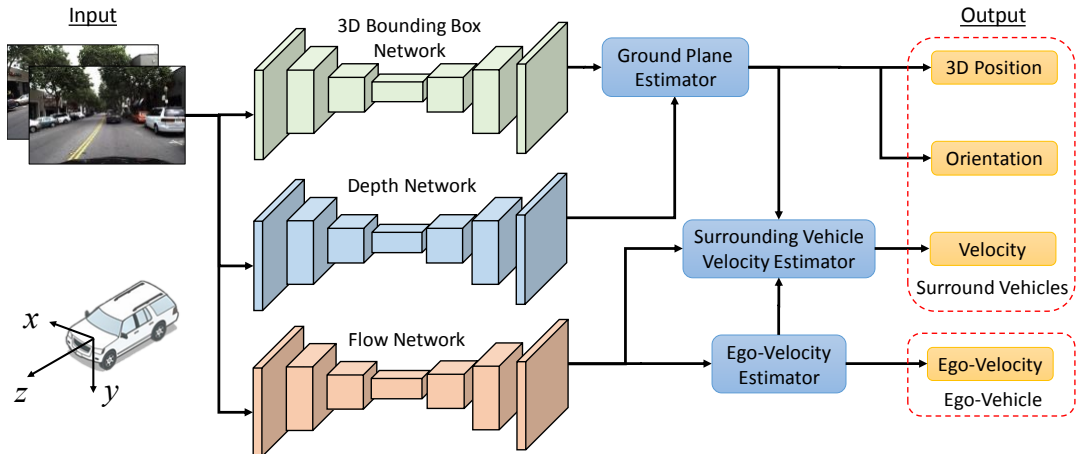

Figure 1. Algorithm overview. To predict 3D position, orientation and velocity of surround vehicles, we combine three deep neural networks. . Our approach is focusing on Ground Plane Estimator, Surrounding Vehicle Velocity Estimator and Ego-Velocity Estimator for calculating the ego-vehicle and surround vehicle state.

processing is not required. If measurements of the ground plane parameters are available, [14] proposes a different approach for ground plane parameter estimation and tracking by the stereo camera. This method is based entirely on the stereo disparity map. Additionally, the approach uses nine points in the lower half of the image for calculating the 3D position of the road surface to improve robustness and stabilize the ground plane estimates. The reported results indicate excellent estimation in real-time.

In order to complement monocular camera data, depth estimation is often used. The approach presented in [10] predicts depth using a model trained by supervised learning. While their method has been shown to work well when training and test data are drawn from the same distribution, generalization and adaptation to new domains cannot be guaranteed if data has a different distribution than the training data.

Other unsupervised methods use the depth estimates directly from stereo images [11][12]. Since training the neural network makes use of the stereo pair output, these algorithms can generate a good depth map even for evaluation images that are different from those used in training. Subsequently, the ground plane parameters can be estimated using the depth map instead of the disparity map from the stereo camera.

In order to estimate ego-motion and surrounding vehicle velocity, optical flow is often used in computer vision because it can be used to calculate the velocity relative to the ground. Traditional optical flow estimation algorithms are categorized into the two classes: feature-based and variational approaches. The feature based method finds image displacements by tracking features such as edges, corners, and other well localized structures and tracks them across a sequence of frames. A major limitation of this method is that it is difficult to estimate flow in regions that lack prominent features, such as the road surface. The variational approach offers a more accurate estimate by coupling the brightness constancy and spatial smoothness assumptions using an energy function. However, this approach is computationally expensive for solving complex optimization problems.

Recently, learning-based optical flow estimation by CNNs have shown significant performance improvements over traditional methods. One such method is FlowNet2.0 [15]. In subsequent research, PWC-Net [16] achieved better performance, a 17 times reduction in in model size, and less complex training than Flownet2.0 model using pyramidal processing, warping, and the use of a cost volume.

## II. FEATURE EXTRACTION FROM MONOCULAR CAMERA

This section presents a new method for estimating ego-motion and surround vehicle state estimation. The framework takes as input an image sequence and combines three state of the art deep neural networks that estimate the 3D vehicle bounding box, depth, and optical flow in conjunction with ground plane estimation to produce estimates of ego-motion and surround vehicle state. An overview of the proposed framework is illustrated in Figure 1.

### A. 3D bounding box estimation

To detect the 3D bounding box of a surrounding vehicle, we use the algorithm proposed in [9] combining DenseBox and a multi-scale network inspired SSD and MS-CNN. This algorithm uses a novel 3D bounding box representation which is independent of the image projection matrix, making it is possible to easily learn with a variety of datasets as training input. This method can output the 2D image coordinates of three bottom vertices and height of the bounding box. Each bottom vertex coordinate $[u\ \nu]^T$ can be reconstructed in the 3D world using the projection matrix (1), (2) and the ground plane equation (3).

$$P = \begin{bmatrix} f_x & 0 & c_x \\ 0 & f_y & c_y \\ 0 & 0 & 1 \end{bmatrix} \begin{bmatrix} r_{11} & r_{12} & r_{13} & t_1 \\ r_{21} & r_{22} & r_{23} & t_2 \\ r_{31} & r_{32} & r_{33} & t_3 \end{bmatrix} \quad (1)$$

$$\begin{bmatrix} x \\ y \\ z \\ 1 \end{bmatrix} = P^{-1} \begin{bmatrix} u \\ \nu \\ 1 \end{bmatrix} \quad (2)$$

where $f$ is the focal length, $c$ is the camera center, $r$ is the rotation factor, and $t$ is the translation factor. The 3D position $[x\ y\ z]^T$ represents the 3D coordinate of each bottom vertex of the 3D bounding box. The 3D coordinates for the top vertices are estimated using the height output. Finally, we can estimate the 3D bounding box as well as the position and orientation of the vehicles in the 3D world coordinate system.

We use a model trained on the KITTI dataset [17] for predicting the 3D bounding box. However, this method is highly dependent on ground plane coefficients. Specifically, the prediction accuracy in the longitudinal (Z-axis in Figure 1) direction changes considerably with these coefficients. Therefore, for accurate 3D estimation of the position and orientation of the bounding box, it is necessary to update the ground plane coefficients by measuring the actual inclination of the ground in real-time.

### B. Ground plane correction

Ground plane estimation is one of the significant factors affecting the estimation accuracy of the 3D position and orientation of surrounding vehicles. Ground plane coefficients are assumed to be determined initially. We use the RANSAC (Random Sample Consensus) algorithm [18] for fitting to the optimal coefficients $(a,b,c,d)$ in (3) using the four bottom corner vertices of the ground truth 3D bounding box in advance.

$$ax + by + cz + d = 0 \tag{3}$$

However, while the ego-vehicle is moving, ground plane coefficients continuously change, depending on the inclination of the road surface and pitching, rolling, and yaw angle of the ego-vehicle. Therefore, ground plane coefficients should be fit to the road surface in real-time. In order to solve this problem, we propose an approach similar to the method presented in [14] for the ground plane correction that assumes the presence of the road surface in the lower part of the image. Whereas the approach in [14] uses a stereo disparity map, our approach estimates depth using a monocular camera. Details of the depth estimation algorithm described in section C. Since the depth calculated by the depth estimator is normalized, we convert the depth to the actual distance based on the known distance.

$$Dist = k \times Depth \tag{4}$$

where $k$ is the coefficient between the depth information ($Depth$) and actual distance ($Dist$).

We choose nine fixed points in the lower portion of the image with the traffic road surface. Subsequently, several points that are inappropriate for estimating the road surface are removed using the detected 3D bounding boxes for every frame. The remaining fixed points are then projected onto a 3D world coordinate system using depth information generated by the depth estimator. Because of its calculation speed and robustness, we then use the RANSAC algorithm, instead of IRLS (Iteratively Reweighted Least-Squares) and LMedS (Least Median of Squares Regression), to estimate the corrected ground plane coefficients. To determine whether an update on the ground plane is necessary, we compare the initial ground plane coefficients, coefficients in the previous frame, and currently estimated coefficients (5). Then we compute the new coefficients of the ground plane (6).

$$\begin{cases} \|n_t - n_{init}\| < \theta_0, \\ \|n_t - n_{t-1}\| < \theta_1, \\ |d_t - d_{init}|/d_{init} < \theta_2 \end{cases} \tag{5}$$

$$a_{new}x + b_{new}y + c_{new}z + d_{new} = 0 \tag{6}$$

where $n_t$ is the normal vector and $d_t$ is the $d$ coefficient of the ground plane for $t^{th}$ frame, $n_{init}$ and $d_{init}$ are the normal vector and $d_{init}$ coefficient of the primary ground plane. Also, $\theta$ is the thresholds for determining whether it is necessary to update the ground plane.

### C. Depth estimation

For depth estimation from a single image, deep neural networks have been shown to be effective for estimating pixel-level depth [12]. This method can be adopted for ground plane estimation, especially for predicting the depth of nine points on the road surface. We use a model pre-trained on the KITTI dataset [17]. Furthermore, since this method uses unsupervised learning, it is possible to ignore influences such as variation of the ground truth annotation among annotators, and it is possible to eliminate the cost of the annotation.

As we mentioned in Section B, we use depth information instead of a stereo disparity map for ground plane correction. The precise depth information of the fixed nine points on the road surface helps with the ground plane correction.

### D. Optical flow estimation

To realize autonomous and ADAS functions, accurate estimation of the surround vehicle's state is necessary. While it is possible to calculate the differential 3D position of the surrounding vehicles using the estimated 3D bounding box detection, this method only provides the relative velocity. Estimation of the absolute and relative velocity of the surround vehicles is an important but challenging problem from a monocular camera.

Optical flow can provide estimates of the absolute velocity of the ego-vehicle based on the flow of stationary objects such as the ground. The absolute velocity of surrounding vehicles can then be estimated using both the relative velocity of the surrounding vehicles and the absolute velocity of the ego-vehicle. In this paper, we use PWC-Net [16] for predicting optical flow with a model pre-trained on KITTI dataset [17].

## III. EGO-MOTION AND SURROUND VEHICLE STATE ESTIMATION

### A. Ego-vehicle velocity estimation

To estimate the velocity of the ego-vehicle, a road surface region near the ego-vehicle is used to calculate the ego-vehicle. This region is depicted by the red-rectangle in Figure 2. The 2D flow vector $(u,v)$ in this fixed rectangular area is then computed and extracted into a 3D flow $(flow_{Gx}, flow_{Gy}, flow_{Gz})$ using the projection matrix (1), (2) and the ground plane (3). Subsequently, the ground speed $V_G$ at the fixed calculation area is then computed using the following relationship,

$$V_G(V_{Gx}, V_{Gy}, V_{Gz}) = \frac{1}{n}\sum_{i=1}^{n} flow_G^i(x,y,z) \quad (7)$$

where $-V_{Gz}$ and $-V_{Gx}$ represent the longitudinal and lateral ego-velocity on the fixed rectangular area, respectively, and $n$ is the total pixel number in the fixed rectangular area. Since the flow is projected on the ground plane surface, the ground speeds in the longitudinal (Z-axis in Figure 1) and lateral (X-axis in Figure 1) directions are predictable. The ground speed is needed for correcting the velocity of the surrounding vehicles.

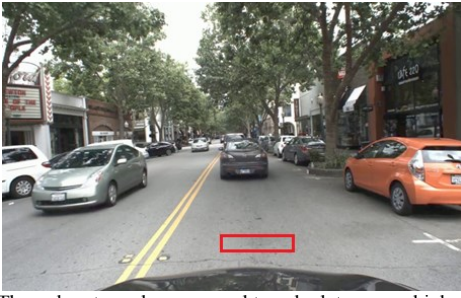

Figure 2. The red rectangular area used to calculate ego-vehicle velocity.

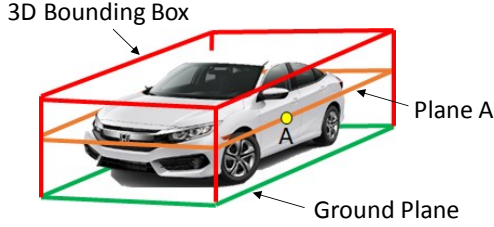

Figure 3. Extraction of the 2D flow based on the detected 3D bounding box. Plane A indicates a plane parallel to the ground plane and passing through an arbitrary point A.

### B. Surrounding Vehicle Velocity Estimation

To estimate the surrounding vehicle velocity, the 2D image flow of the detected vehicles must be projected to the corresponding 3D coordinates of the vehicles.

The 3D bounding box position of the surrounding vehicles together with coordinates of their vertices is estimated based on the formulation in section II.A. Given that an arbitrary 2D point in the detected 3D bounding box can be converted into the 3D position, a new plane is passing through the arbitrary point and parallel to the ground plane can be calculated at each point on the detected 3D bounding boxes in Figure 3. We can convert the 2D flow into a 3D flow based on the new plane corresponding to the surrounding vehicles. For example, the 2D flow at the A point in Figure 3 has to be projected on "Plane A" not on the ground plane.

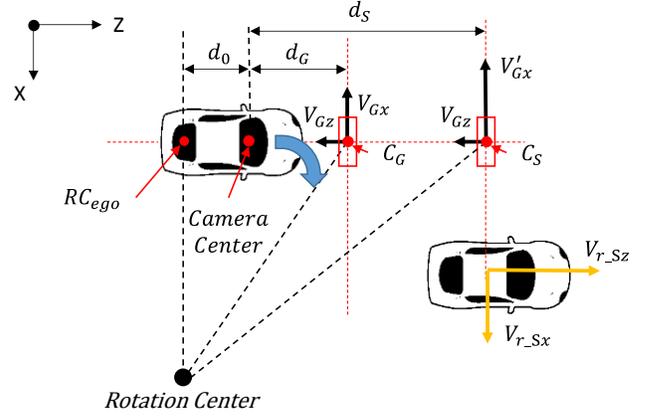

Figure 4. Geometry diagram.

Furthermore, as we showed in Figure 3, the rectangular parallelepiped 3D bounding box has not been precisely fitted to the shape of the vehicle. Particularly in the upper half of the 3D bounding box, there is a large variation in the vehicle shape, depending on the vehicle type (e.g., sedan, SUV, or hatchback). Therefore, in our approach, we calculate the flow in the lower half of the 3D bounding box. Moreover, we use two vertical planes near the ego-vehicle in the detected 3D bounding box for 2D flow extraction because these two vertical planes can be visible from the ego-vehicle if there are no occlusions by other obstacles. The 2D flow vector $(u,v)$ in the 3D bounding box can be projected on each plane based on the 2D coordinates. We can then compute the 3D flow and the relative velocity of the surrounding vehicles using the following:

$$V_{r\_S}(V_{r\_Gx}, V_{r\_Gy}, V_{r\_Gz}) = \frac{1}{m}\sum_{i=1}^{m} flow_G^i(x,y,z) \quad (8)$$

where $m$ is total pixel number in the lower half of the 3D bounding box. Since $V_{r\_S}$ is the relative velocity, conversion to absolute velocity is necessary.

The calculation of absolute velocity differs between the longitudinal and lateral direction. Initially, concerning the longitudinal (Z-axis) velocity, the absolute longitudinal velocity $V_{a\_Sz}$ can be calculated by the relative velocity of the surrounding vehicle $V_{r\_Sz}$ and the ground velocity $V_{Gz}$ in Figure 4 as follows:

$$V_{a\_Sz} = V_{r\_Sz} + V_{Gz} \quad (9)$$

Regarding lateral (X-axis) velocity, we need to recalculate a ground velocity $V'_{Gx}$ near the surrounding vehicle from the calculated ground velocity $(V_{Gx}, V_{Gz})$ in the fixed area in Figure 4. The distance $d_0$ between $RC_{ego}$ and the camera center is constant but depends on the vehicle type. The distance $d_G$ between the camera center and the center $C_G$ of the fixed ground plane, and the distance $d_S$ between the camera center and the center $C_S$ of the ground near the surrounding vehicle can be calculated from the 3D coordinates of the fixed rectangular area and the detected surrounding vehicles. The ground velocity $V'_{Gx}$ near the surrounding vehicle can be calculated in (10). The absolute lateral velocity $V_{a\_Sx}$ can be calculated in (11).

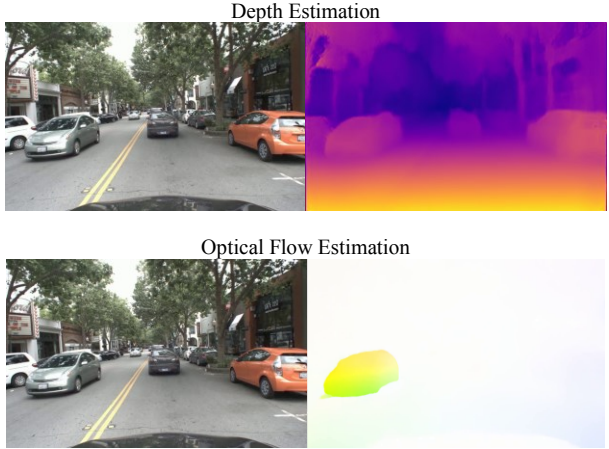

Figure 5. Depth and optical flow estimation results.

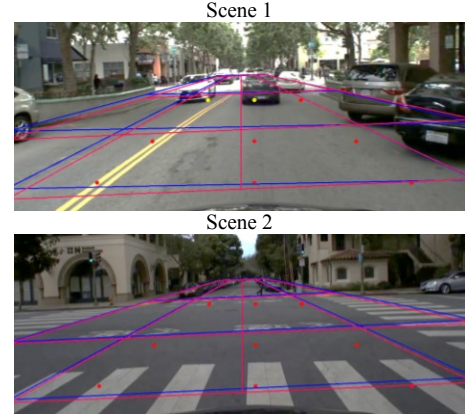

Figure 6. Ground plane correction results. Red and yellow dots represent the fixed nine points on the road surface. The yellow dots are excluded for the ground plane correction by the vehicle detection results. The red and blue lines represent the initial and corrected ground plane estimates, respectively.

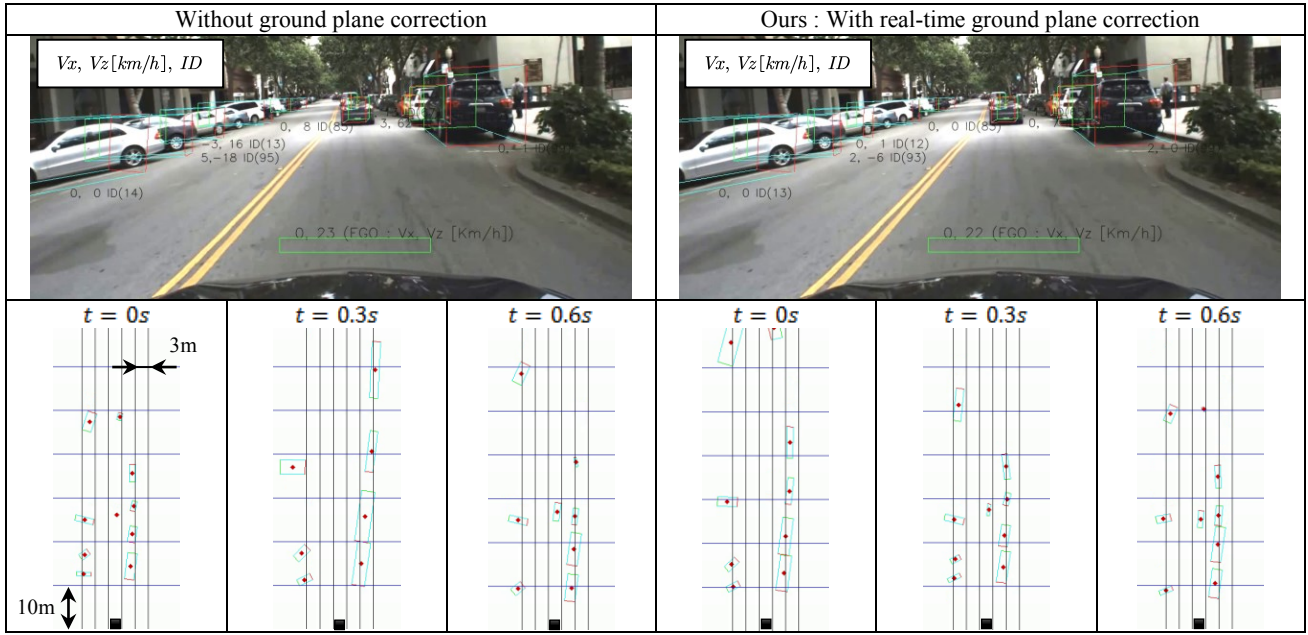

Figure 7. 3D position comparison results without ground plane correction (left) and with real-time ground plane correction (right). The vehicle positions, orientations, and bounding box size are more stable after correction.

$$V'_{Gx} = \frac{(d_0 + d_S)}{(d_0 + d_G)} \times V_{Gx} \quad (10)$$

$$V_{a\_Sx} = V_{r\_Sx} + V'_{Gx} \quad (11)$$

The results with absolute longitudinal and lateral velocity are available for understanding more detailed surrounding vehicle's state.

*C. Surround vehicle position and orientation estimation*

The position and orientation of surrounding vehicles can be calculated by projecting the detected 3D bounding box into the corrected ground plane. The corrected ground plane estimates using depth information in real-time provide accurate and robust estimates for position and orientation.

IV. EXPERIMENTAL RESULTS

*A. Settings*

We use our evaluation dataset with monocular camera images in front of the vehicle and LiDAR, GPS and CAN-BUS information. This information is synchronized. Our evaluation data were captured in Palo Alto, CA USA and also include the ego-vehicle velocity from CAN-BUS data.

| | Lateral direction distance error [m] | | | | Longitudinal direction distance error [m] | | | |
|---|---|---|---|---|---|---|---|---|
| | MAE | RMSE | iMAE | iRMSE | MAE | RMSE | iMAE | iRMSE |
| Without correction | 0.55 | 1.24 | 0.79 | 2.89 | 1.92 | 2.70 | 0.009 | 0.013 |
| Ours : With correction | **0.52** | **1.19** | **0.53** | **2.17** | **1.24** | **1.70** | **0.006** | **0.009** |

Table 1. Quantitative comparison for 3D position estimation between using real-time ground plane correction and without ground plane correction. (50 evaluation samples)

| | Average Orientation Similarity (AOS) Scores [%] |
|---|---|
| Without correction | 97.15 |
| Ours : With correction | **98.97** |

Table 2. Quantitative comparison for orientation estimation between using real-time ground plane correction and without ground plane correction. (50 evaluation samples)

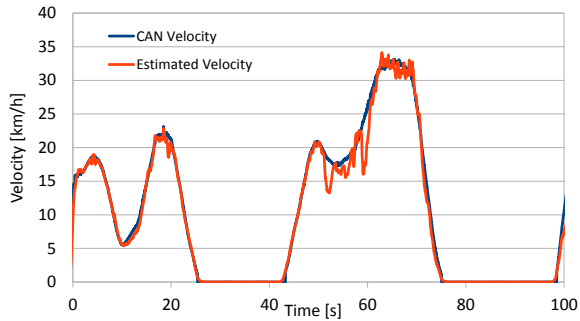

Figure 8. Ego-vehicle velocity estimation result

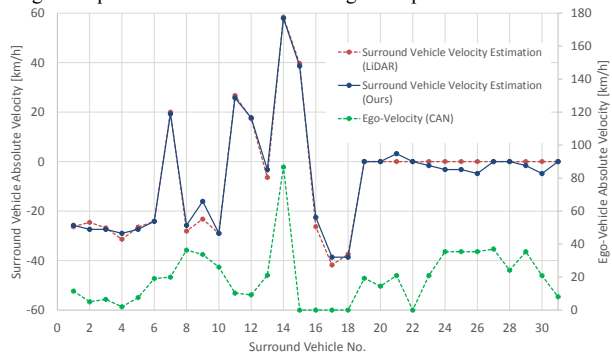

Figure 9. Surround vehicle absolute velocity estimation result

| Ego-vehicle [$km/h$] | | Surrounding vehicle [$km/h$] | |
|---|---|---|---|
| MAE (ego) | RMSE (ego) | MAE (surround) | RMSE (surround) |
| 0.56 | 1.18 | 1.74 | 2.49 |

Table 3. Absolute velocity estimation error of ego-vehicle and surrounding vehicle. (31 evaluation samples)

The Input camera image resolution is $1280 \times 800$ at 30 fps, the fixed calculation area for the ground velocity is $320 \times 30$ section III.A, the coefficient $k = 0.13$ in (4), the thresholds are $\theta_0 = 0.025$, $\theta_1 = 0.004$, $\theta_2 = 0.02$ in (5), $d_0 = 2.0m$ in Figure 4.

*B. Results*

Estimation results of depth and optical flow are shown in Figure 5. Qualitatively, the estimated depth and flow correlate well with the depth and flow in our dataset using the model trained with the KITTI dataset. We estimate the ground plane using depth information at nine points on the road surface in real-time. The real-time ground plane correction results are shown in Figure 6. To validate these results, we compared the 3D position and orientation estimation results of the surrounding vehicles with and without real-time ground plane correction using our evaluation data. Comparison results are shown in Figure 7, Table 1, and Table 2. Without ground plane correction, the 3D position, orientation, and bounding box size of the surrounding vehicles abnormally fluctuate due to the change of the ground surface. With ground plane correction, the 3D position of the vehicles is more stable and accurate. These results indicate that ground plane correction reduces 3D position estimation errors attributed to changes in the ground surface or optical axis.

Ego-vehicle and surround vehicle velocity estimation results are reported in Figure 8, Figure 9, and Table 3. The ego-vehicle velocities estimation results are compared to the ground truth obtained from the vehicle CAN-BUS data as shown in Figure 8 and Table 3. The Root Mean Square Error (RMSE) of the estimated ego-velocity is $1.18\ km/h$, which is sufficient accuracy in our applications. Comparison of the estimated velocity of the surrounding vehicle with LiDAR estimation in our evaluation dataset is shown in Figure 9 and Table 3. We evaluate the velocity of thirty-one surrounding vehicles for several different scenes. The RMSE of the estimated velocity of surrounding vehicle is $2.49\ km/h$, which is sufficient accuracy for understanding surrounding vehicle velocity.

V. CONCLUSION

We presented a new method to estimate ego-motion and surrounding vehicle state using a monocular camera. Our method can accurately estimate the 3D position, velocity, and orientation of surrounding vehicles using calculated depth and optical flow for real-time ground plane correction and ego-velocity. In addition, our method is suitable for combination with future, novel algorithms of depth, optical flow, and 3D bounding box estimation.